\documentclass{article}
\usepackage{iclr2024_conference,times}

\usepackage{hyperref}
\usepackage{graphicx}	
\usepackage{tablefootnote}
\usepackage{color, soul}
\usepackage{caption}
\usepackage{amsmath}
\usepackage{subcaption}
\usepackage{hyperref}
\usepackage{bbm}
\usepackage{booktabs}
\usepackage{siunitx}

\setcitestyle{authoryear,round,citesep={;},aysep={,},yysep={;}}

\iclrfinalcopy 
\begin{document}

\title{Imbalance-aware Presence-only Loss Function for Species Distribution Modeling}

\author{Robin Zbinden, Nina van Tiel \\
EPFL, Switerzland \\
\texttt{firstname.lastname@epfl.ch} 
\And  Marc Rußwurm\\
WUR, Netherlands \\
\texttt{marc.russwurm@wur.nl} 
 \AND Devis Tuia\\
EPFL, Switerzland \\
\texttt{devis.tuia@epfl.ch} 
}

\maketitle

\begin{abstract}
In the face of significant biodiversity decline, species distribution models (SDMs) are essential for understanding the impact of climate change on species habitats by connecting environmental conditions to species occurrences. Traditionally limited by a scarcity of species observations, these models have significantly improved in performance through the integration of larger datasets provided by citizen science initiatives. However, they still suffer from the strong class imbalance between species within these datasets, often resulting in the penalization of rare species--those most critical for conservation efforts. To tackle this issue, this study assesses the effectiveness of training deep learning models using a balanced presence-only loss function on large citizen science-based datasets. We demonstrate that this imbalance-aware loss function outperforms traditional loss functions across various datasets and tasks, particularly in accurately modeling rare species with limited observations.
\end{abstract}

\section{Introduction}


Species distribution models (SDMs) play a crucial role in ecology, serving as indispensable tools for understanding and predicting the spatial distribution of species. By establishing correlations between species occurrence data and environmental variables~\citep{elith2009species}, these models provide valuable insights into the ecological niches and habitat preferences of diverse organisms, thereby informing conservation efforts~\citep{guisan2013predicting}. The significance of SDMs in identifying and safeguarding endangered species becomes even more pronounced as habitats of numerous species face imminent threats from climate change~\citep{thomas2004extinction, dyderski2018much}. Additionally, conservation endeavors aimed at preventing biodiversity loss not only contribute to mitigating climate change~\citep{shin2022actions} but also play an important role in alleviating its broader impacts~\citep{portner2023overcoming}.

These conservation efforts primarily focus on rare and endangered species, which are inherently difficult to observe. Consequently, we have only a few observations available, posing challenges to the development of reliable SDMs~\citep{breiner2015overcoming}.
Recent initiatives in citizen science present promising avenues to facilitate the collection of large amounts of species records. However, there is still a high disparity in the number of observations per species, ranging from a few handfuls of occurrences to tens of thousands for the most common or iconic species~\citep{botella2023geolifeclef2023, cole2023spatial}. Such a significant \textit{class imbalance} reflects the existence of various biases within the data, which can be geographical and taxonomic, among others~\citep{feldman2021trends}. Additionally, the data gathered through citizen science initiatives is typically \textit{presence-only}, i.e., it consists of recorded occurrences but no data regarding the species' absence~\citep{pearce2006modelling}. Managing such data limitations brings additional complexities in developing accurate and reliable SDMs for rare species.

\begin{figure}[t]
    \begin{subfigure}[t]{0.485\textwidth}
    \includegraphics[width=1\linewidth]{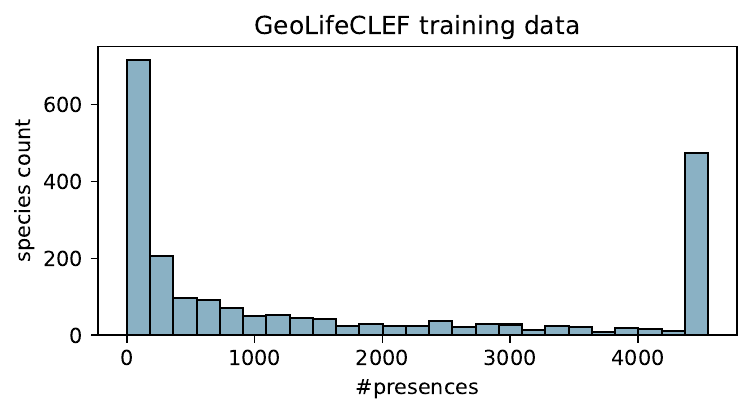} 
    \vspace{-20pt}
    \label{fig:longtail_geolifeclef}
    \end{subfigure}
    \begin{subfigure}[t]{0.5\textwidth}
    \includegraphics[width=1\linewidth]{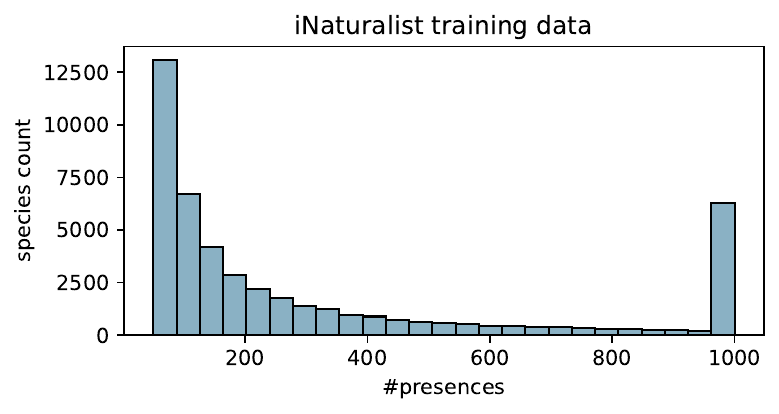}
    \vspace{-20pt}
    \label{fig:longtail_inat}
    \end{subfigure}
    
    \caption{Distributions of the number of presence records in the GeoLifeCLEF 2023 (left) and iNaturalist (right) training datasets, obtained through citizen science initiatives. Both distributions exhibit a long-tailed pattern, which is crucial to address to avoid penalizing rare species during training.}
    \label{fig:longtail}
\end{figure}

Deep learning (DL) has demonstrated promise for SDMs~\citep{deneu2021convolutional, teng2023bird}, by enabling the simultaneous modeling of multiple species and the identification of shared environmental patterns, a characteristic particularly advantageous for rare\footnote{In this study, for simplicity, we refer to rare species as those with the lowest numbers of occurrences in the training set. It is important to note that the rarity of species is a more complex concept, dependent on factors such as range size, occupancy, and abundance~\citep{crisfield2024and}.} species~\citep{zbinden2023species}. However, the persistence of high class imbalance in training datasets (see Figure~\ref{fig:longtail}) often leads to the neglect of rare species during model training, despite their critical importance. Recently,~\cite{zbinden2024selection} introduced a balanced loss function to account for the class imbalance between species, demonstrating its effectiveness in improving the model's performance on rare species, but only on the relatively small datasets of~\cite{elith2020presence}. In this study, we extend their work to larger datasets acquired through citizen science initiatives at continental and global scales~\citep{botella2023geolifeclef2023, cole2023spatial}. Specifically, we train DL models with different presence-only losses and evaluate these models on different SDM-related tasks. 
Our findings highlight that an imbalance-aware loss function is essential to achieve optimal performance on rare species. 

\section{Methods}

Generally, SDMs learn correlations between the environmental features and observed species occurrence patterns. Depending on the downstream application, the model can also be made \textit{spatially explicit}~\citep{domisch2019spatially} by incorporating geospatial coordinates as additional input through location encoders~\citep{russwurm2023geographic}. Multi-species distribution models aim to learn the distribution of multiple species simultaneously within a single model.
The task then involves multi-label classification, as a given location may host an arbitrary number of species. However, a significant portion of species records exists in the form of presence-only data, often with only one species presence observation associated with a given location. This creates a scenario known as \textit{single positive multi-label learning}~\citep{cole2021multilabel}, which poses significant challenges.
To cope with this scenario, a common strategy involves sampling \textit{pseudo-absences} (PAs), designating samples as negative even when certainty about the absence of a species is lacking, and incorporating them into the loss function during model training~\citep{cole2023spatial}. Below, we describe such losses.

\subsection{Loss functions}
\label{sec:losses}

The predominant approach for multi-label classification uses the binary cross-entropy loss function~\citep{nam2014large, ung2023leverage}. For a given location with species observations represented by $y$, it is defined as follows:
\begin{align}
    \mathcal{L}_{\text{BCE}}(\mathbf{y}, \mathbf{\hat{y}}) = - \frac{1}{S} \sum^S_{s=1}\bigl[ \mathbbm{1}_{[y_s=1]} \log(\hat{y}_s) + \mathbbm{1}_{[y_s=0]} \log(1 - \hat{y}_s) \bigr]
\end{align}
where $y_s$ is \num{1} if species $s$ has been observed and \num{0} otherwise, $\hat{y}_s \in [0,1]$ is the predicted suitability score for species $s$, and $S$ denotes the number of species. It is essential to note that $y_s = 0$ doesn't necessarily imply that the species is absent; it simply indicates that it hasn't been observed. This corresponds to a PA, characterized here by the observation of another species. These specific PAs are referred to as target-group background points in the context of SDMs~\citep{phillips2009sample}. Since target-group background points are only located where other species' observations are, they may not entirely cover the area of interest. 
To address this limitation,~\citet{cole2023spatial} extended the BCE loss to incorporate, for each sample, a PA located at a random location with a predicted score $\hat{y}_s'$, introducing the \textit{full assume negative loss} function:
\begin{align}
    \mathcal{L}_{\text{full}}(\mathbf{y}, \mathbf{\hat{y}}) = - \frac{1}{S} \sum^S_{s=1}\bigl[ \mathbbm{1}_{[y_s=1]} \lambda \log(\hat{y}_s) + \mathbbm{1}_{[y_s=0]} \log(1 - \hat{y}_s) + \log(1 - \hat{y}_s') \bigr].
\end{align}
Here, $\lambda=2048$ is included to counterbalance the larger number of PAs compared to presences. However, this loss doesn't address the high class imbalance between species, which is particularly detrimental for rare species. To tackle this issue, we introduced class-specific weights, or species weights, in a prior work~\citep{zbinden2024selection}, resulting in the \textit{full weighted loss} function:
\begin{align}
    \mathcal{L}_{\text{full-weighted}}(\mathbf{y}, \mathbf{\hat{y}}) = - \frac{1}{S} \sum^S_{s=1} \bigl[ \mathbbm{1}_{[y_s=1]} \lambda_1 w_s  \log(\hat{y}_s)   +  \mathbbm{1}_{[y_s=0]} \lambda_2 \frac{1}{\left(1 - \frac{1}{w_s}\right)}  \log(1 - \hat{y}_s) \nonumber  \\ + (1 - \lambda_2) \log(1 - \hat{y}_s')\bigr].
\end{align}
Species weights are defined as $w_s = \frac{n}{n_{\text{p($s$)}}} = \frac{1}{\text{freq(\textit{s})}}$,
where $n_{\text{p($s$)}}$ denotes the number of presences of species $s$, and $n$ represents the total number of presence locations, i.e., the number of training samples. Additionally, the weight $\lambda_2$ is introduced to modulate the impact of different types of PAs. In this work, we extensively test these loss functions across different large-scale citizen science-based datasets and tasks, with a particular emphasis on rare species. The datasets are described in the following section, and training details are presented in the Appendix~\ref{sec:training_details}.

\subsection{Datasets}
\label{sec:datasets}

\textbf{GeoLifeCLEF 2023 (GLC23).} Originally designed for a competition~\citep{botella2023geolifeclef2023}, this dataset provides an extensive collection of species observation records from various citizen science sources. GLC23 includes a public validation set containing presence-absence data for \num{2174} plant species. We use it to assess performances by calculating the mean AUC across all species. Additionally, we compute a mean AUC for \num{468} rare species (with \num{50} observations or fewer). The distribution of the number of presence records per species is shown in the left panel of Figure~\ref{fig:longtail}.

\textbf{iNaturalist.} We leverage the codebase, model architectures, datasets, and tasks developed in~\cite{cole2023spatial}. Specifically, the training dataset comprises \num{35.5} million observations spanning \num{47375} species from iNaturalist\footnote{\url{https://www.inaturalist.org/}}. Unlike the GLC23 dataset, all species in this dataset have a minimum of \num{50} observations, with some species exceeding \num{100000} occurrences. For computational efficiency, we limit the number of observations to \num{1000} per species (see right panel of Figure~\ref{fig:longtail}), aligning to~\cite{cole2023spatial}.
We evaluate with the following three tasks: first, the eBird Status and Trends (\textbf{S\&T}) test set \citep{fink2020ebird}, derived from expert range maps, which includes \num{535} bird species. The number of presences for these species is less imbalanced, with almost all the species having a substantial number of observations during training. Specifically, we categorize the \num{96} out of \num{535} species with less than \num{1000} occurrences as rare. Second, the International Union for Conservation of Nature (\textbf{IUCN}) test set, which is more imbalanced and contains \num{639} rare species with \num{100} occurrences or less. Performance is evaluated using the mean average precision (mAP) for both the S\&T and IUCN test sets. Third, the models are assessed as \textbf{geographic priors} for fine-grained image classification. This task enhances species image classification by incorporating location and environmental metadata into the model. We calculate the top-1 accuracy gain ($\Delta$ Top-1) by adding SDMs to complement the vision model. The distribution of the number of presences of the species considered in these three tasks is presented in the Appendix~\ref{sec:test_sets}.

\section{Results}

\setlength{\tabcolsep}{9pt} 
\renewcommand{\arraystretch}{1.4} 
\begin{table}[t]
\centering
\resizebox{\columnwidth}{!}{%
\begin{tabular}{lccccccc}
    \toprule 
   & \multicolumn{2}{c}{\textbf{GeoLifeCLEF}} & \multicolumn{2}{c}{\textbf{S\&T}} & \multicolumn{2}{c}{\textbf{IUCN}} & \textbf{Geo Prior}  \\
   & \multicolumn{2}{c}{(AUC)} &
   \multicolumn{2}{c}{(mAP)}         & \multicolumn{2}{c}{(mAP)}         & ($\Delta$ Top-1)                  \\
   & all                  & rare        & all                  & rare        & all                     & rare          & all                       
   \\ 
   \cmidrule(lr){1-1}\cmidrule(lr){2-3}\cmidrule(lr){4-5}\cmidrule(lr){6-7}\cmidrule(lr){8-8}  
$\mathcal{L}_{\text{BCE}}$                                  & 0.453               & 0.398 & \textbf{0.810}                & 0.746      & 0.702                & 0.659       & +6.7                                            \\
$\mathcal{L}_{\text{full}}$                                  & 0.786               & 0.802 & 0.807                & \textbf{0.756}       & \underline{0.761}                & \underline{0.703}       & +6.6                                             \\ 
\cmidrule(lr){1-1}
$\mathcal{L}_{\text{full-weighted}}$ with $ \lambda_2 = 0.5$ & \textbf{0.796}            & \textbf{0.854} & 0.806                & 0.751       & \textbf{0.765}       & \textbf{0.710}       & \textbf{+7.3}                                    \\
$\mathcal{L}_{\text{full-weighted}}$ with $ \lambda_2 = 0.8$ & \underline{0.787}      & \underline{0.841}   & \textbf{0.810}       & \underline{0.753}       & 0.757                & 0.700       & \textbf{+7.3}                                 \\ \bottomrule
\end{tabular}%
}
\caption{Performance of the different losses function on the SDMs tasks. Results in bold correspond to the best in the column, while the second-best is underlined.}
\label{tab:results}
\end{table}

\begin{figure}[t]
    \begin{subfigure}[t]{0.5\textwidth}
    \includegraphics[width=1\linewidth]{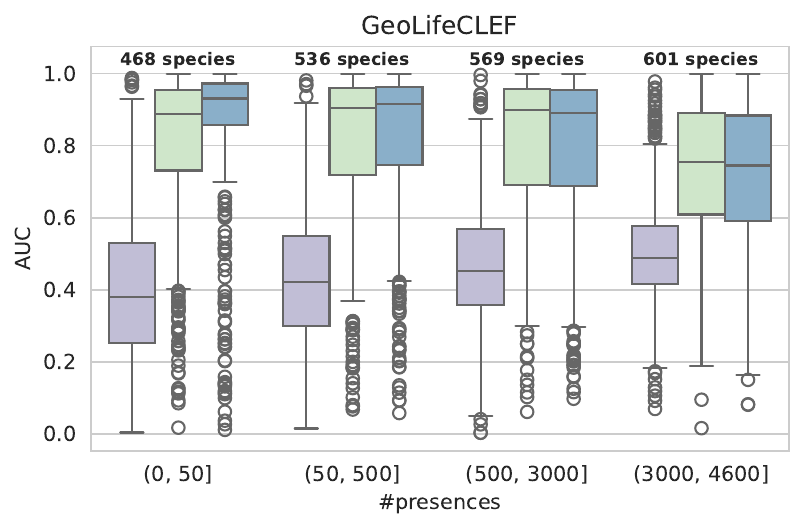} 
    \vspace{-20pt}
    \label{fig:results_geolifeclef}
    \end{subfigure}
    \begin{subfigure}[t]{0.5\textwidth}
    \includegraphics[width=1\linewidth]{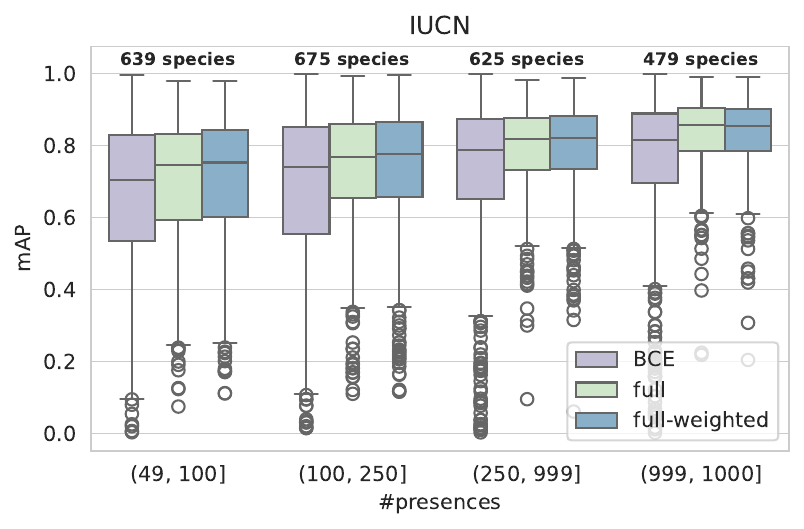}
    \vspace{-20pt}
    \label{fig:results_iucn}
    
    \end{subfigure}
    
    \caption{Performance of the loss functions, grouped by the number of presences records of species in the training set. The $\mathcal{L}_{\text{full-weighted}}$ loss, defined here with $\lambda_2 = 0.5$, is beneficial for rare species.}
    \label{fig:results}
\end{figure}

Results are presented in Table~\ref{tab:results} and Figure~\ref{fig:results}. Firstly, we observe consistent high performance with the full weighted loss, specifically with $\lambda_2=0.5$, outperforming the other losses in three out of four tasks. Notably, substantial improvements are observed in the GLC23 dataset, particularly for rare species. This aligns well with the nature of the dataset, which contains numerous rare species with a limited number of observations. As depicted in the left panel of Figure~\ref{fig:results}, the $\mathcal{L}_{\text{full-weighted}}$ loss demonstrates improved performance over the $\mathcal{L}_{\text{full}}$ loss as the number of presences in the training set decreases.

Results are more nuanced for the S\&T dataset, where all loss functions yield similar performance. This may be attributed to the fact that most bird species have at least \num{1000} observations in the training set, as illustrated in Figures~\ref{fig:longtail_all} and~\ref{fig:results_snt} in the Appendix, which diminishes the effect of the full weighted loss.
In contrast, the IUCN dataset presents more variability, leading to a slightly higher performance with the $\mathcal{L}_{\text{full-weighted}}$ loss. Additionally, we note the significantly higher performance on the Geo Prior task, potentially due to the test set being more balanced than the training set. Since this task involves multi-class classification, it favors balanced loss functions. 
Finally, the $\mathcal{L}_{\text{full-weighted}}$ loss with $\lambda_2 = 0.5$ seems to perform slightly better than its counterpart with $\lambda_2 = 0.8$.

\section{Conclusion}

In this study, we emphasized the importance of effectively modeling the distribution of rare species using deep learning, which requires addressing the high class imbalance commonly found in datasets derived from citizen science initiatives.
The presented results illustrate the advantages of employing a balanced loss function for SDMs across three out of four datasets, demonstrating substantial performance improvements for rare species. Notably, achieving equal performance on the S\&T dataset, despite its lower imbalance, suggests that imbalance-aware loss functions do not adversely affect less imbalanced applications.
Lastly, we stress the significance of considering species with very few observations in benchmark datasets when evaluating SDMs, as is done in the GeoLifeCLEF 2023 dataset. This aspect is particularly crucial given that SDMs are most valuable when aimed at predicting the distribution of rare and endangered species.

\bibliography{literature}

\newpage

\appendix

\section{Training details}
\label{sec:training_details}

\textbf{GeoLifeCLEF 2023 (GLC23).}
Our models are trained on the \num{2856818} presence records corresponding to the \num{2174} species in the test set. 
We provide the models with \num{19} historical bioclimatic variables, \num{9} pedological variables, and \num{17} land cover classes as tabular data. We employ a multi-layer perceptron model (MLP) with \num{5} fully connected hidden layers, each containing \num{1000} neurons and connected through residual connections~\citep{gorishniy2021revisiting}. Batch normalization~\citep{ioffe2015batch} is applied to each hidden layer, and the training process spans $150$ epochs with an SGD optimizer and a learning rate set to $0.001$. Finally, the $\mathcal{L}_{\text{full-weighted}}$ loss function uses $\lambda_1 = 1$.

\textbf{iNaturalist.}
We adopt the identical configuration and MLP architecture as described in~\cite{cole2023spatial}. We focus on their approach that incorporates both coordinates and environmental data as inputs. For the $\mathcal{L}_{\text{full-weighted}}$ loss function, we set $\lambda_1 = 0.1$ since the inverse of the frequency of species presences becomes very large with the high number of species considered during training.

\section{Distributions of species presences}
\label{sec:test_sets}

\begin{figure}[h]
    \centering
    \begin{subfigure}[t]{0.46\textwidth}
        \includegraphics[width=\linewidth]{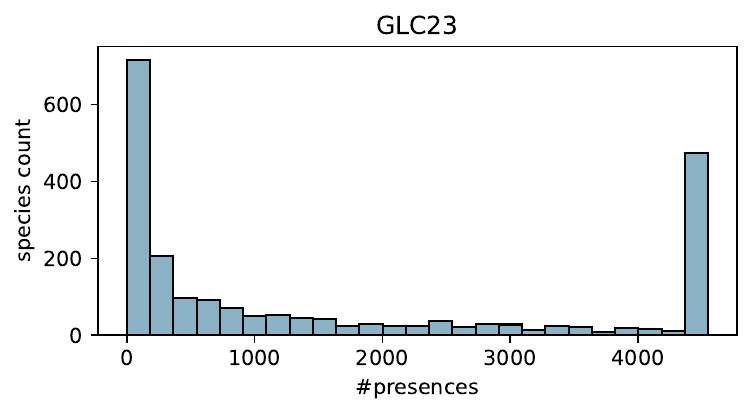} 
        \label{fig:species_counts_geolifeclef}
    \end{subfigure}
    \hfill
    \begin{subfigure}[t]{0.46\textwidth}
        \includegraphics[width=\linewidth]{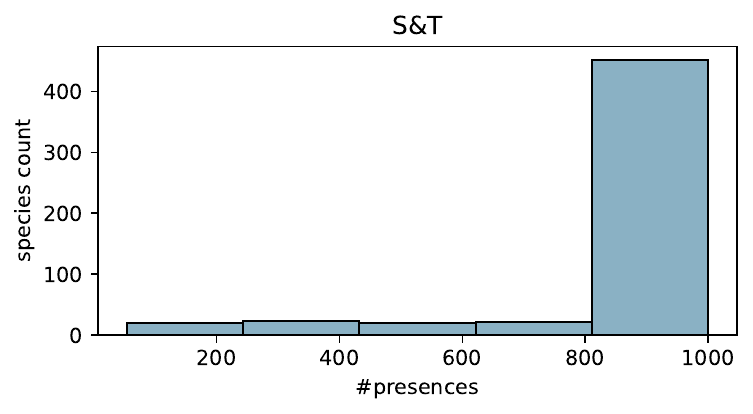} 
        \label{fig:species_counts_snt}
    \end{subfigure}
    
    \vspace{-10pt}
    
    \begin{subfigure}[t]{0.46\textwidth}
        \includegraphics[width=\linewidth]{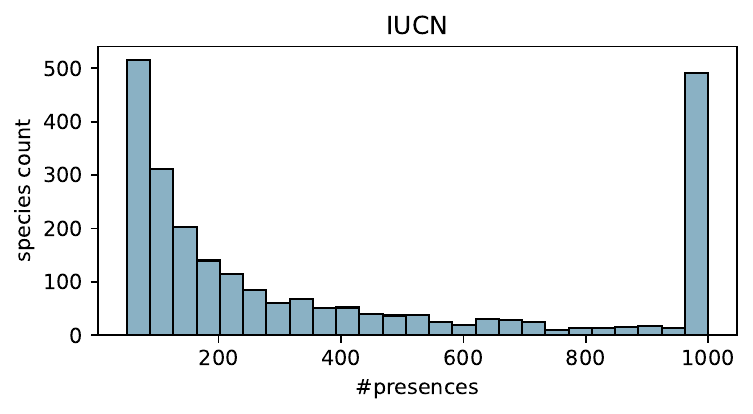}
        \label{fig:species_counts_iucn}
    \end{subfigure}
    \hfill
    \begin{subfigure}[t]{0.47\textwidth}
        \includegraphics[width=\linewidth]{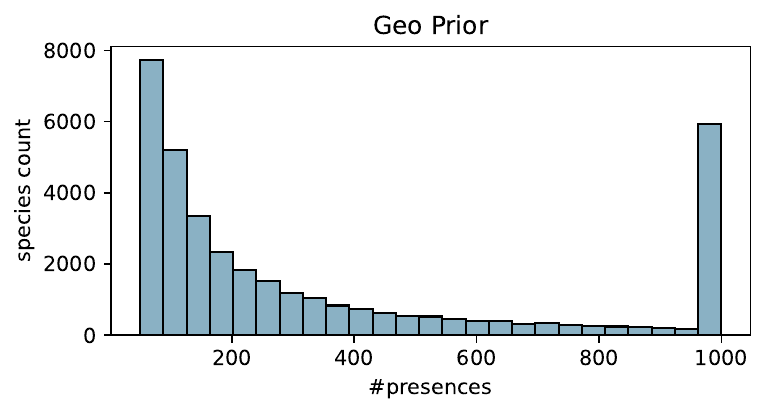}
        \label{fig:species_counts_geoprior}
    \end{subfigure}

    \vspace{-13pt}
    
    \caption{Distribution of the number of training presences of the species considered in the different tasks. The GLC23 training set contains the same species used in testing.}
    \label{fig:longtail_all}
\end{figure}

\section{Extra S\&t results}

\begin{figure}[h]
    \centering
    \includegraphics[width=0.5\linewidth]{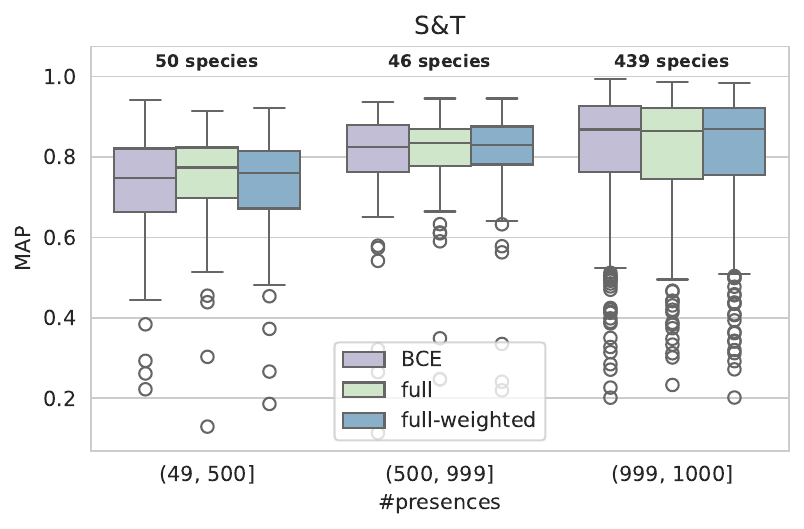}
    \caption{Performance of the loss functions on the S\&T dataset, grouped by the number of presences records of species in the training set.} 
    \label{fig:results_snt}
\end{figure}

\end{document}